\documentclass[5p, final, times, authoryear]{elsarticle}

\usepackage{lineno,hyperref}
\usepackage{url}
\usepackage{booktabs} 
\usepackage{textcomp}
\usepackage{xcolor}
\usepackage{amsmath,amssymb,amsfonts}
\usepackage{algorithmic}
\usepackage{graphicx}
\usepackage{subcaption}

\title{Binarized Simplicial Convolutional Neural Networks}

\begin{document}
\author[1]{Yi Yan
}
\ead{y-yan20@mails.tsinghua.edu.cn, yiyiyiyan@outlook.com}
\author[1]{Ercan Engin Kuruoglu\corref{cor1}
}
\ead{kuruoglu@sz.tsinghua.edu.cn}

\address[1]{Tsinghua Shenzhen International Graduate School, Tsinghua University, Shenzhen, China.}
\cortext[cor1]{Corresponding author. This work was supported by the Tsinghua Shenzhen International Graduate School Start-up Fund under Grant QD2022024C and Shenzhen Science and Technology Innovation Commission
under Grant JCYJ20220530143002005.}

\begin{abstract}
Graph Neural Networks have the limitation of processing features solely on graph nodes, neglecting data on high-dimensional structures such as edges and triangles. Simplicial Convolutional Neural Networks (SCNN) represent high-order structures using simplicial complexes to break this limitation but still lack time efficiency. In this paper, a novel neural network architecture named Binarized Simplicial Convolutional Neural Networks (Bi-SCNN) is proposed based on the combination of simplicial convolution with a weighted binary-sign forward propagation strategy. The utilization of the Hodge Laplacian on a weighted binary-sign forward propagation enables Bi-SCNN to efficiently and effectively represent simplicial features with higher-order structures, surpassing the capabilities of traditional graph node representations. The Bi-SCNN achieves reduced model complexity compared to previous SSCN variants through binarization and normalization, also serving as intrinsic nonlinearities of Bi-SCNN; this enables Bi-SCNN to shorten the execution time without compromising prediction performance and makes Bi-SCNN less prone to over-smoothing. Experimenting with real-world citation and ocean-drifter data confirmed that our proposed Bi-SCNN is efficient and accurate. 
\end{abstract}
\maketitle

\section{Introduction}
Graphs and networks are gaining enormous awareness in the field of machine learning and artificial intelligence owing to their ability to effectively learn and represent irregular data interactions \citep{Ortega_2018, Leus_2023_Graph, Sandryhaila_2014_big_data}. 
Graph structures can be deduced from various real-life applications by embedding the features onto the nodes of graph representations using the graph Laplacian. 
Sensor graphs of nationwide temperature \citep{bib_NLMS} and air quality \citep{bib_GCN_air} are built by combining sensor readings with the sensor location. 
Citation graphs \citep{McCallum2000automating} can be formed by linking the citation relationship between authors.
The brain region activation can be modeled using a series of dynamic graphs \citep{zhao_2024_sequential}.
Graphs and topological structures can be used to model the stock market \citep{2024_Stable_Muvunza, 2024_Market_Qin}.
In the past, spatial and spectral graph neural networks (GNNs) such as the Graph Convolutional Networks (GCN) \citep{defferrard2016convolutional, kipf2016semi}, Graph Attention Networks (GAT) \citep{velivckovic2017graph}, and their variants have had major success on graph representation learning tasks including node classification, graph classification, and link prediction. 
The GCN and the GAT can be generalized by the Message Passing Neural Networks (MPNN) where instead of Global level graph representation done by the graph Laplacian matrix or the graph adjacency matrix, a localized message-passing scheme is defined for each node  \citep{fey_2019_fast}. 
However, GNNs operate only on the features residing on the graph nodes and do not pay enough attention to the graph edges and the structures formed by the edges. 
In the real world, data or features can appear not only on graph nodes but also on high-order structures.
Traffic flow could be recorded on either the edges \citep{Schaub_2018} or the nodes \citep{yu2017spatio} of a graph. 
A map of ocean drifter trajectories can be modeled using the data on graph edges \citep{Schaub_2020_Random}. 
In a citation complex, a collaboration among three or even more authors is modeled using simplicial complexes, which is a scenario that cannot be simply recorded using the graph nodes only because a traditional representation of an edge connection between two nodes can only represent a one-to-one relationship between two authors \citep{ebli2020_SNN}. 
More such examples where data appears on structures with high-order than graph nodes include pandemic prediction based on population mobility among regions \citep{Panagopoulos2021pandemic} and dynamic gene interaction graphs \citep{Kuruoglu_2016}. 

Recently the concept of Topological Signal Processing (TSP) was introduced: instead of the graph Laplacian, a more generalized Laplacian known as the Hodge Laplacian is adopted in TSP to incorporate the representation of high-order features that are beyond the graph nodes  \citep{Barbarossa_2020}. 
In TSP, operations such as spectral filtering, spatial diffusion, and convolution are defined on simplicial complexes using the Hodge Laplacians \citep{Yang_2022_Simplicial}. 
High-dimensional feature representations such as data on the graph edges and data on the triangles formed by three edges can be embedded into simplicial complexes using methodologies analogous to graph (node) embedding \citep{martino2019hyper}. 

If we feed the outputs of the TSP into non-linear active functions and concatenate a few of these together, we can transform them into neural networks. 
The Simplicial Neural Network (SNN) uplifts the GCN onto simplicial complexes by using TSP to extend the convolutional ability of GCN to process high-order simplicial features \citep{ebli2020_SNN}.
The Simplicial Convolutional Neural Networks (SCNN) \citep{Yang_2022_SCNN} further improves the performance of SNN by applying the Hodge decomposition to the simplicial convolution so it is able to train the upper and lower simplicial adjacencies separately. 
The Simplicial Attention Network (SAT) \citep{goh2022_SAT} and the Simplicial Attention Neural Network \citep{Giusti_2022_SAN} introduced the attention mechanism to SNN and SCNN to process simplicial features, making them generalized versions of the GAT from graphs to simplicial complexes.  
The Simplicial Message Passing Networks (Simplicial MPNN) have made the simplicial complex analogy of defining message passing on graphs \citep{fey_2019_fast} to simplicial complexes \citep{bodnar_2021_weisfeiler}. 
The Dist2Cycle was proposed to use Simplicial complexes to define neural networks for Homology Localization \citep{keros_2022_dist2cycle}.
These architectures have shown to be successful at applications such as imputing missing data on the citation complexes of different data dimensions and classifying different types of ocean drifter trajectories.  

Although the simplicial complex-based neural networks have achieved numerous successes, the focus of the previous networks was on how to extend concepts such as convolution or attention from graphs onto simplicial complexes, which leads to two drawbacks.
First, the recently proposed simplicial neural network structures do not prioritize time efficiency, resulting in inefficiency in terms of computational complexity and run time. 
Let us look into the GCN, which is the 0-simplex analogy of the SNN, the Simplified GCN \citep{wu2019_SimplifiedGCN} has proved a trained GCN essentially is a low-pass filter in the spectral domain and the complexity of GCN can be reduced by eliminating nonlinearities and collapsing weights. 
The Binary-GCN (Bi-GCN) \citep{Wang2021_Bi_GCN} was inspired by the Binary Neural Networks to use binary operations such as XOR and bit count to speed up training time. 
In the field of Graph Signal Processing, the Adaptive Graph-Sign algorithm derived from $l_1$-optimization results in a simple yet effective adaptive sign-error update that has outstanding run speed and robustness \citep{yan_2022_sign, yan_2023_sign}.  
To the best of our knowledge, similar improvements that focus on the efficiency and complexity of neural network algorithms on the simplicial complex are absent. 
Following the first drawback of the current neural network algorithms on simplicial complexes, the high complexity of current simplicial neural network structures may cause the resulting simplicial embeddings to get over-smoothed in some scenarios. 
In their graph predecessors, the over-smoothed phenomenon was extensively analyzed \citep{li2018deeper, liu2020_over_smooth}, with solutions such as spectral filtering \citep{zhu2021simple} and regularizations\citep{1013Chen_2023_AGNN} to alleviate over-smoothing, but has not yet gained enough attention in the simplicial complex-based neural networks. 

In this paper, we propose a novel neural network architecture on simplicial complexes named Binarized Simplicial Convolutional Neural Networks (Bi-SCNN) with the following advantages:  
\begin{itemize}
\item The TSP foundation of the Bi-SCNN enables it to effectively represent and process data residing on higher-order simplicial complexes, which extends its learning capabilities beyond just operating on graph nodes.
\item The Bi-SCNN can be interpreted using spatial and spectral filters defined using the TSP backbone.
\item The combination of simplicial convolution with a weighted binary-sign propagation strategy significantly reduces the model complexity and the training time of Bi-SCNN compared to existing algorithms such as SCNN, without compromising the prediction accuracy. 
Moreover, the Bi-SCNN inherently incorporates nonlinearity and activation functions through the use of the Sign() function and feature normalization.
\item  Bi-SCNN is less prone to over-smoothing as a result of the efficient architectural complexity. 
\end{itemize}

This paper is organized as follows. 
Section~\ref{sec_related_work} provides a discussion of the related work. 
The preliminary knowledge is discussed in Section~\ref{sec_preliminary}.
The derivation of the Bi-SCNN is presented in Section~\ref{sec_method}. The experiment results and related discussion can be found in Section~\ref{sec_experiment}. 
Section~\ref{sec_conclusion} concludes the paper.
\section{Related Works}
\label{sec_related_work}

Algorithms that conduct convolution on the nodes of a graph such as GCN \citep{kipf2016semi} operate only on the graph nodes, but not on $k$-simplexes with $k\geq1$. 
Both the SNN \citep{ebli2020_SNN} and the SCNN \citep{Yang_2022_SCNN} are direct extensions of the GCN onto simplicial complexes; SNN is architecturally identical to the SCNN except SNN does not use Hodge decomposition. 
The Bi-SCNN is an extension of the SCNN \citep{Yang_2022_SCNN}, but the Bi-SCNN uses feature normalization and feature binarization which the SCNN does not.
Also, the weighted binary-sign forward propagation strategy in the Bi-SCNN layer has 2 outputs but the SCNN layer only has 1 output, making a neural network that contains more than 2 layers of Bi-SCNN a drastically different architecture than using purely SCNN. 

These convolutions on graphs or simplicial complexes are enabled using aggregations defined by Laplacian or adjacency matrices where the weights are assigned to the global topology level. 
To increase the representation power of convolutions on graphs or simplicial complexes, message passing based on lower and upper adjacencies can be defined on the local level \citep{fey_2019_fast, bodnar_2021_weisfeiler}. 
On graphs, the MPNN defines trainable weights on all of the edges between the nodes instead of only a small number of weights on the graph (normalized or unnormalized) Laplacian as seen in the GCN \citep{fey_2019_fast}.
On simplicial complexes, a more powerful but complicated message-passing scheme can be defined using the lower, upper, boundary, and coboundary adjacencies \citep{bodnar_2021_weisfeiler}.
MPNNs indeed boost the representational power of GNNs or SNNs, but this advantage comes with higher memory demands, longer training periods, scalability challenges, and difficulties with over-smoothing due to an increased number of trainable parameters \citep{Giovanni_2023_MPNN, jiang2024limiting}.

The Bi-GCN \citep{Wang2021_Bi_GCN} has an acronym similar to the Bi-SCNN, but we should emphasize that the Bi-SCNN is not simply a simplicial complex version of the Bi-GCN and they are drastically different from each other with the following four differences. 
Firstly, the Bi-GCN operates only on the graph nodes, while the Bi-SCNN operates on $k$-simplexes. 
Secondly, as explained earlier, the Bi-SCNN does not follow the traditional GCN/SCNN forward propagation strategy because the feature binarization and the feature normalization propagate separately. 
Thirdly, the Bi-GCN binarizes both the feature and the trainable weights but Bi-SCNN only binarizes the features. 
Fourthly, the Bi-GCN uses binary operations bit count and XOR to speed up the original GCN forward propagation and uses predefined values to deal with the backpropagation of binary operations; the Bi-SCNN uses a weighted binary-sign forward propagation, separating the pass for feature normalization and feature binarization in both forward and backward propagation. 

\begin{figure}
    \centering
    \includegraphics[trim={180 350 180 350},clip,width=0.4\textwidth]{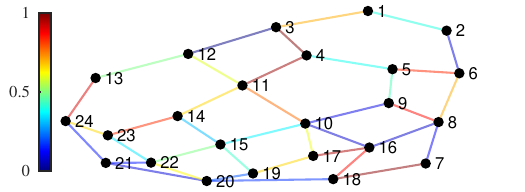}
    =\caption{A graph with $N_0 = 24$ nodes, $N_1 = 38$ edges, and $N_2 = 2$ triangles having signals on the edges.}
    \label{fig_edge_signal}
\end{figure}

\section{Preliminaries}
\label{sec_preliminary}
We will begin with a set of vertex $\mathcal{V} = {\{v_0 ... v_{N_0}}\}$ to define a $k$-simplex $\mathcal{S}_k$, where $\mathcal{S}_k$ is a subset of $\mathcal{V}$ with cardinality $k+1$. 
Throughout the paper, subscript $k$ is used to denote which $\mathcal{S}_k$ a variable belongs to.
The collection of $k$-simplexes with $k = 0 ... K$ is a simplicial complex $\mathcal{K}$ of order $K$. 
If generalizing a graph $\mathcal{G} = \{ \mathcal{V, E} \}$ as a simplicial complex, the vertex set $\mathcal{V}$ of  $\mathcal{G}$ is the node set that belongs to $\mathcal{S}_0$ and the edge set $\mathcal{E} = \{ e_1 ... e_{N_1} \}$ belongs to $\mathcal{S}_1$. 
Note that for a simplex $\mathcal{S}_k \in \mathcal{K}$, if $\mathcal{S}_{k-1}$ is a subset of $\mathcal{S}_k$, then $\mathcal{S}_{k-1} \in \mathcal{K}$. 
We use the variable ${N_k}$ to denote the number of elements of $\mathcal{S}_k$.
Take a graph $\mathcal{G}$ as an example, ${N_0} = |\mathcal{V}|$ is the number of nodes and ${N_1} = |\mathcal{E}|$ is the number of edges. 
Each simplicial feature vector of a simplex $\mathcal{S}_k$ is a mapping from the $k$-simplex to $\mathbb{R}^{N_k}$ and will be denoted as $\boldsymbol{x}_k = [x_{k,0} ... x_{k,N_k}]^T$ \citep{Barbarossa_2020} . 
Simplicial data with $d$ features will be denoted as $\mathbf{X}_k$ = [$\boldsymbol{x}_{k,0} ... \boldsymbol{x}_{k,d}$]. 
A graph containing only signals on the edges and defined using the above definition is shown in Figure~\ref{fig_edge_signal}. 

To process the data on a simplicial complex, we will be using the Hodge Laplacian matrix defined as follows: 
\begin{equation}
    \mathbf{L}_k = 
    \begin{cases}
    \mathbf{B}_{k+1}\mathbf{B}_{k+1}^T, & k = 0,\\
        \mathbf{B}_k^T\mathbf{B}_k+\mathbf{B}_{k+1}\mathbf{B}_{k+1}^T = \mathbf{L}_{k,l}+\mathbf{L}_{k,u},  
        & 0 < k < K,\\
            \mathbf{B}_{k}^T\mathbf{B}_{k}, &  k = K,
    \end{cases}
    \label{hodge_laplacian}
\end{equation}
where the $\mathbf{B}_k \in \mathbb{R}^{N_{k-1} \times N_{k-1}}$ is the incidence matrix that represents the relationship between adjacent simplices $\mathcal{S}_{k-1}$ and $\mathcal{S}_{k}$ in $\mathcal{K}$. 
Each row of $\mathbf{B}_k$ correspond to one element in $\mathcal{S}_{k}$, and each column of $\mathbf{B}_k$ correspond to one element in $\mathcal{S}_{k-1}$. 
Each adjacency relationship between $\mathcal{S}_{k-1}$ and $\mathcal{S}_{k}$ will result in a $|1|$ entry in $\mathbf{B}_k$ with the sign determined by the orientation of the element in $\mathcal{S}_{k}$. 
One useful property of the incidence matrix is $0 = \mathbf{B}_k\mathbf{B}_{k+1}$.

One can see an important characteristic of $\mathbf{L}_k$, for $1 \leq k<K$, $\mathbf{L}_k$ can be split into $\mathbf{L}_{k,l}$ and $\mathbf{L}_{k,u}$ \eqref{hodge_laplacian}; $\mathbf{L}_{k,l}$ is the lower Hodge Laplacian that represents the lower adjacencies of $\mathcal{S}_k$, and $\mathbf{L}_{k,u}$ is the upper Hodge Laplacian that represents the upper-adjacency of $\mathcal{S}_k$ \citep{Barbarossa_2020}. 
It is worth pointing out an important property of the simplicial Fourier transform (SFT) in the spectral domain: the eigenvectors paired with the non-zero eigenvalues of $\mathbf{L}_{k,u}$ are orthogonal to the eigenvectors paired with the non-zero eigenvalues of $\mathbf{L}_{k,l}$ \citep{Barbarossa_2020}.  
The graph Laplacian matrix seen in GCN or GAT is $\mathbf{L}_0$. 
Two examples of incidence matrices are $\mathbf{B}_1$ and $\mathbf{B}_2$, which are the node-to-edge incidence matrix and the edge-to-triangle incidence matrix respectively.

\section{Binarized Simplicial Convolutional Neural Networks}
\label{sec_method}
We begin with a discussion of spectral and spatial TSP techniques.
Then, the simplicial convolution on a $k$-simplex, which is the theoretical foundation of our Bi-SCNN, will be analyzed using TSP.
The advantages and drawbacks of several simplicial-convolution-based algorithms are discussed and analyzed.
Afterward, we will provide the derivations of Bi-SCNN and combine feature normalization, binary-sign updates, simplicial filters, and nonlinearities to form our Bi-SCNN layer. 
Lastly, we will give a detailed interpretation of the Bi-SCNN.
The goal is to improve the simplicial convolution shown in \eqref{conv3} by reducing its computational complexity and making it less prone to over-smoothing. 

\subsection{Simplicial Convolution and Topological Signal Processing}

In short, the simplicial convolution is essentially the conducting the following procedure: 1. transforming spatial data into the spectral domain, 2. applying a filter in the spectral domain to the simplicial data, 3. using the inverse transform to transform the filtered data back into the spatial domain \citep{Barbarossa_2020}. 
This simple yet efficient procedure is made possible due to the convolution property, which can be summarized as convolution becomes multiplication in the frequency domain. 
In GNNs, operations such as graph convolution and embedding of the node features $\mathbf{X}_0$ are possible by the computations defined by the graph Laplacian. 
For instance, the spatial graph convolution in GCN aggregates the data from the 1-hop neighbors of each node by $\mathbf{L}_0\mathbf{X}_0$ \citep{kipf2016semi}. 
From the Graph Signal Processing perspective, the graph convolution can also be defined by applying low-pass filters in the spectral domain to data on the graph nodes \citep{defferrard2016convolutional}. 
In order to define these spectral domain operations on the graph nodes, the node features are viewed as signals and are transformed to the spectral domain using the Graph Fourier Transform. 
The Graph Fourier Transform is the graph (node) analogy of the Fourier Transform found in classical Signal Processing \citep{Ortega_2018}. 
In classical signal processing, signals are projected onto the spectral domain on a Fourier basis. 
In the case where signals appear on the graph nodes, the eigenvectors of the graph Laplacian matrix $\mathbf{L}_0$ serve as the Fourier basis and the corresponding eigenvalues are the frequencies.
When the Graph Fourier Transform analogy is made, lower eigenvalues and their corresponding eigenvectors are referred to as the low-frequency components.
Intuitively, in the low-frequency components, each signal on the nodes will have similar values to its neighbors; in other words, the low-frequency node signals have smoother variations over the graph topology.
On the other hand, higher eigenvalues and their corresponding eigenvectors are the high-frequency components, which represent the signals that change more rapidly over the graph.

\begin{figure}[h]
     \centering
     \begin{subfigure}[b]{0.4\textwidth}
         \centering
         \includegraphics[trim={180 350 180 350},clip,width=\textwidth]{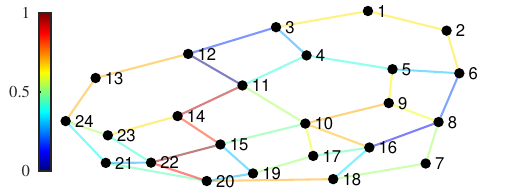}
         \caption{$\boldsymbol{u}_1$}
     \end{subfigure}
     \hfill
     \begin{subfigure}[b]{0.4\textwidth}
         \centering
         \includegraphics[trim={180 350 180 350},clip,width=\textwidth]{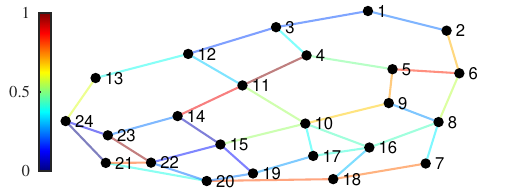}
         \caption{$\boldsymbol{u}_{20}$}
\end{subfigure}
     \hfill
     \begin{subfigure}[b]{0.4\textwidth}
         \centering
         \includegraphics[trim={180 350 180 350},clip,width=\textwidth]{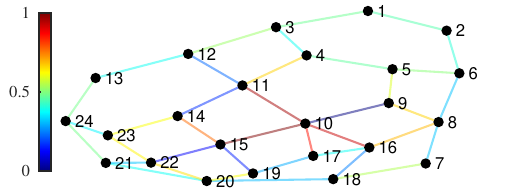}
         \caption{$\boldsymbol{u}_{N_1}$}
\end{subfigure}
        \caption{Decomposing the edges of the graph with $N_1 = 38$ edges in Figure~\ref{fig_edge_signal} using the SFT in \eqref{eq_SFT} and $k=1$.}
    \label{fig_SFT}
\end{figure}

In TSP, the SFT is the analogy of the classical Fourier transform, which enabled the definition of spectral operations on simplicial complexes.
However, to represent high-order structures, the Hodge Laplacian will be used to define the transform instead of using only the graph Laplacian. 
For a set of features to be able to transform between domains in a $k$-simplex, we need the SFT defined by the eigendecomposition of the Hodge Laplacian matrix. 
For the $k$-simplex, the SFT is the eigendecomposition \citep{Barbarossa_2020}:
\begin{equation}
    \mathbf{L}_k = \mathbf{U}_k\mathbf{\Lambda}_k\mathbf{U}_k^T,
    \label{eq_SFT}
\end{equation}
where $\mathbf{U}_k$ is the eigenvector matrix and $\mathbf{\Lambda}_k$ is the eigenvalue matrix.
Following the graph Fourier transform convention, the eigenvalues are again assigned the notion of frequency. 
The eigenvalues on the diagonals of $\mathbf{V}_k$ are arranged in increasing order.
Each eigenvector $\boldsymbol{u}_{i = 1 ... N_1} $is the column of $\mathbf{U}_k$ and is arranged to appear in the same order as the eigenvalues.
A few visual illustrations of the spectral components obtained from the SFT are shown in Figure~\ref{fig_SFT}.

The forward SFT projects a simplicial signal $\boldsymbol{x}_k$ into the spectral domain, thereby representing the signal in terms of frequency components:
\begin{equation}
    \tilde{\boldsymbol{x}}_k = \mathbf{U}_k^T\boldsymbol{x}_k,
\end{equation} 
where $\tilde{\boldsymbol{x}}_k$ is $\boldsymbol{x}_k$ transformed into the spectral domain.
Similarly, the inverse SFT will invert the process of the forward SFT in which the signal $\boldsymbol{x}_k$ is being reconstructed from the spectral domain back to the spatial domain:
\begin{equation}
    \boldsymbol{x}_k = \mathbf{U}_k\tilde{\boldsymbol{x}}_k.
\end{equation}

\begin{figure}[h]
     \centering
     \begin{subfigure}[b]{0.4\textwidth}
         \centering
         \includegraphics[trim={180 350 180 350},clip,width=\textwidth]{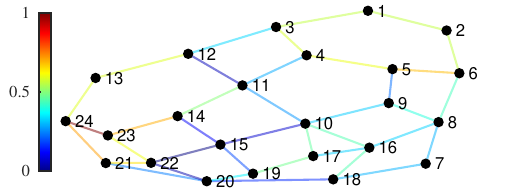}
         \caption{The edge signal in Figure~\ref{fig_edge_signal} going through a low-pass filter.}
     \end{subfigure}
     \hfill
     \begin{subfigure}[b]{0.4\textwidth}
         \centering
         \includegraphics[trim={180 350 180 350},clip,width=\textwidth]{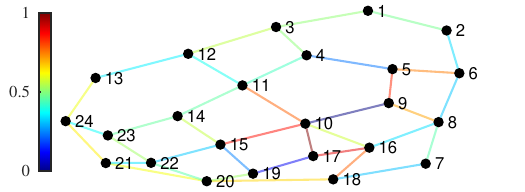}
         \caption{The edge signal in Figure~\ref{fig_edge_signal} going through a high-pass filter.}     
\end{subfigure}
        \caption{An illustration of simplicial filtering on an edge signal using \eqref{conv1}.}
    \label{fig_SFT_filter}
\end{figure}

As we pointed out earlier, from the signal processing perspective, the modification of the signal will be conducted in the spectral domain, which requires the definition of a simplicial filter. 
A simplicial filter can be represented using a function on the frequencies matrix $h(\mathbf{\Lambda}_k)$. 
Afterward, using the convolution property, applying this filter to the simplicial signal $\boldsymbol{x}_k$ can be achieved in the spectral domain by
\begin{equation}
\boldsymbol{y}_k = \mathbf{U}_kh(\mathbf{\Lambda}_k) \tilde{\boldsymbol{x}}_k =\mathbf{U}_k h(\mathbf{\Lambda}_k) \mathbf{U}_k^T {\boldsymbol{x}}_k,
\label{conv1}
\end{equation}
where $\boldsymbol{y}_k$ is the result of the simplicial convolution.
For an input simplicial signal $\boldsymbol{x}_k$, different filters such as the low-pass filters or the high-pass filters can be applied using \eqref{conv1} to enhance desired features, suppress unwanted noise, or isolate specific components. 
Figure~\ref{fig_SFT_filter} provides an illustration of applying low-pass and high-pass filters to the edge signal $\boldsymbol{x}_1$.

The spectral convolution of a $k$-simplex with feature size $N_k$ in \eqref{conv1} relies on the eigendecomposition that has $O(N_k^3)$ complexity and is numerically unstable when $N_k$ is large.  
Luckily, it has been shown that a series of polynomial summations approximate the above spectral approach, resulting in the spatial simplicial convolution \citep{Yang_2022_Simplicial}:
\begin{equation}
    \mathbf{U}_k h(\mathbf{\Lambda}_k) \mathbf{U}_k^T \boldsymbol{x}_k = h(\mathbf{L}_k){\boldsymbol{x}}_k \approx \sum^J_{j=0} w_{k,j }\mathbf{L}^j_k\boldsymbol{x}_k,
    \label{conv2}
\end{equation}
where $w_{k,j}$ is the filter weight and the $J$ is the filter length of the spatial Simplicial Convolution. 

Let us introduce another concept named the Hodge decomposition before we proceed to the Simplicial Convolution:
\begin{equation}
    \boldsymbol{x} = \boldsymbol{x}_{k, h} + \mathbf{B}_{k}^T\boldsymbol{x}_{k-1}+ \mathbf{B}_{k+1}\boldsymbol{x}_{k+1},
    \label{eq_hodge_decomposition}
\end{equation}
where $\mathbf{B}_{k}^T\boldsymbol{x}_{k-1}$ is the portion of $\boldsymbol{x}_k$ that is induced from $\boldsymbol{x}_{k-1}$, $\mathbf{B}_{k+1}\boldsymbol{x}_{k+1}$ is the portion of $\boldsymbol{x}_k$ that is induced from $\boldsymbol{x}_{k+1}$, and $\boldsymbol{x}_{k,h}$ is the harmonic component that cannot be induced \citep{Barbarossa_2020}. 
The harmonic component $\boldsymbol{x}_{k,h}$ of $\boldsymbol{x}_k$ has the property of
\begin{equation}
    0 = \mathbf{L}_k\boldsymbol{x}_{k,h}.
    \label{harmonic_property}
\end{equation}
In other words, the Hodge decomposition in \eqref{eq_hodge_decomposition} essentially decomposes the Hodge Laplacian matrix $\mathbf{L}_k$ into the gradient, curl, and harmonic components \citep{Yang_2022_Simplicial}.

\subsection{Simplicial Convolution and TSP in Neural Networks}

In \eqref{conv2}, we can assign specific filter weights to the lower and upper Hodge Laplacians because by definition, in \eqref{hodge_laplacian}, the Hodge Laplacian matrix can be split into the upper and the lower Laplacian. 
This leads to the (single feature) convolution operation seen in SCNN \citep{Yang_2022_SCNN}: 
\begin{equation}
    \boldsymbol{y}_k = \sum^{J_l}_{j=1} \gamma_{k,j }\mathbf{L}^j_{k,l}\boldsymbol{x}_k+\sum^{J_u}_{j=1} \theta_{k,j }\mathbf{L}^j_{k,u}\boldsymbol{x}_k+\xi_{k}\mathbf{I}\boldsymbol{x}_k, 
    \label{conv3}
\end{equation}
where $\gamma_{k,j }$ is the weights of the lower Hodge Laplacian and $\theta_{k,j }$ is the weights for the upper Hodge Laplacian. 
A term $\xi_{k}\mathbf{I}\boldsymbol{x}_k$ is added to the simplicial convolution because of the property shown in \eqref{harmonic_property}:  the role of $\xi_{k}\mathbf{I}\boldsymbol{x}_k$ is to take into account of the harmonic component as it can only be processed independently from the upper or lower Laplacian \citep{Barbarossa_2020}. 
Combining the simplicial convolution with nonlinearity, the forward propagation of an SCNN layer is 
\begin{equation}
        \boldsymbol{z}_k^{p+1} =\sigma\left(\sum^{J_l}_{j=1} \gamma_{k,j }\mathbf{L}^j_{k,l}\boldsymbol{z}_k^p+\sum^{J_u}_{j=1} \theta_{k,j }\mathbf{L}^j_{k,u}\boldsymbol{z}_k^p+\xi_{k}\boldsymbol{z}_k^p\right),
        \label{SCNN_layer}
\end{equation}
where $\sigma()$ is the activation function, $p$ is the layer number, and $\boldsymbol{z}_k^p$ is the output of the $p^{th}$ simplicial layer \citep{Yang_2022_SCNN}. 
In the SCNN, $\gamma_{k,j }$, $\theta_{k,j}$, and $\xi_k$ are the trainable weights to be optimized. 
Each $k$-simplex in a simplicial complex with order $K$ will have individually their own $k$-simplex SCNN. 


\subsection{Simplicial Convolution Architectural Analysis}

To facilitate later derivations, we will provide the expression for the multi-variate version of the simplicial convolution in \eqref{conv3} when there are $d$ features in the data $\mathbf{X}_k$:
\begin{equation}
\mathbf{Y}_{k,p+1} = \sum^{J_l}_{j=1} \mathbf{L}_{k,l}^j\mathbf{Z}_{k,p} \mathbf{\Gamma}_{k,j}+\sum^{J_u}_{j=1} \mathbf{L}_{k,u}^j\mathbf{Z}_{k,p}\mathbf{\Theta}_{k,j}+\mathbf{\Xi}_{k}\mathbf{Z}_{k,p},
\label{conv_multi}
\end{equation}
where $\Gamma_{k,j}$, $\Theta_{k,j }$, and $\Xi_{k,j }$ are the trainable weights with their sizes determined by the number of input and output features of layer $p$. 
The SCNN multi-variate version of the SCNN is also provided here as it was not given in the SCNN original literature \citep{Yang_2022_SCNN}: 
\begin{equation}
        \mathbf{Z}_{k,p+1} = \sigma\left(\sum^{J_l}_{j=1} \mathbf{L}_{k,l}^j\mathbf{Z}_{k,p} \mathbf{\Gamma}_{k,j}+\sum^{J_u}_{j=1} \mathbf{L}_{k,u}^j\mathbf{Z}_{k,p}\mathbf{\Theta}_{k,j}+\mathbf{\Xi}_{k}\mathbf{Z}_{k,p}\right).
        \label{eq_SNN_multi}
\end{equation}
In the first layer, the input is $\mathbf{Z}_{k, 0} = \boldsymbol{x}_k$. 
We will refer to the above spatial simplicial convolution as the simplicial convolution throughout the rest of the paper. 

It is worth mentioning that the SCNN reduces to SNN \citep{ebli2020_SNN} if we set $\mathbf{\Xi}_{k} = 0$ and $\mathbf{\Gamma}_{k,j} = \mathbf{\Theta}_{k,j}$. 
This essentially means that the SNN is the SCNN without Hodge decomposition \eqref{eq_hodge_decomposition}.
The multi-variate version of SNN is given as a reference:
\begin{equation}
    \mathbf{Z}_{k,p+1} = \sigma\left(\sum^{J_l}_{j=1} \mathbf{L}_{k}^j\mathbf{Z}_{k,p} \mathbf{\Gamma}_{k,j}\right).
        \label{eq_SCNN_multi}
\end{equation}

For simplicity of analysis, let us assume the length of simplicial convolutions in \eqref{eq_SCNN_multi} is $J_u = J_l = J$. 
Notice that for $J > 1$, we can intuitively acquire each summation term of length $J > 1$ convolution by feeding our input into a series of concatenated $J$ length 1 convolution. 
In other words, one SCNN layer of length $J > 1$ can be represented as the combination of multiple SCNN layers of length $J = 1$ without nonlinearity. 
To see this more clearly, let us consider the case of $J_l = J_u = 2$ and break down \eqref{eq_SCNN_multi}:
\begin{equation}
        \sigma\left(\sum^{2}_{j=1} \mathbf{L}_{k,l}^j\mathbf{Z}_{k,p} \mathbf{\Gamma}_{k,j}+\sum^{2}_{j=1} \mathbf{L}_{k,u}^j\mathbf{Z}_{k,p}\mathbf{\Theta}_{k,j}+\mathbf{\Xi}_{k}\mathbf{Z}_{k,p}\right) = 
        \sigma\left(\mathbf{S}_1+\mathbf{S}_2\right), 
        \label{eq_break}
\end{equation}
where 
\begin{equation}
    \mathbf{S}_1 =\mathbf{L}_{k,l}\mathbf{Z}_{k,p} \mathbf{\Gamma}_{k,1}+\mathbf{L}_{k,u}\mathbf{Z}_{k,p}\mathbf{\Theta}_{k,1}+\mathbf{\Xi}_{k}\mathbf{Z}_{k,p}
\end{equation}
and 
\begin{equation}
    \mathbf{S}_2 = \mathbf{Z}_{k,p}+\mathbf{L}_{k,l}^2\mathbf{Z}_{k,p} \mathbf{\Gamma}_{k,2}+\mathbf{L}_{k,u}^2\mathbf{Z}_{k,p}\mathbf{\Theta}_{k,2}.
    \label{eq_s2}
\end{equation}
It is easy to see that $\mathbf{S}_1$ is just a length 1 convolution, which is equivalent to the case when we set $J = 1$ in \eqref{conv_multi}. 
Now, as for $\mathbf{S}_2$ in \eqref{eq_s2}, it actually can be expressed as 
\begin{equation}
        \mathbf{S}_2 = \mathbf{L}_{k,l}\hat{\mathbf{Y}}_{k,p} \mathbf{\Gamma}_{k,2}+\mathbf{L}_{k,u}\tilde{\mathbf{Y}}_{k,p}\mathbf{\Theta}_{k,2}+0\mathbf{Z}_{k,p} 
\end{equation}
where
\begin{equation}
            \hat{\mathbf{Y}}_{k,p} = \mathbf{L}_{k,l}\mathbf{Z}_{k,p} \mathbf{I}_{k}+\mathbf{L}_{k,u}\mathbf{Z}_{k,p}0+0\mathbf{Z}_{k,p}, 
\end{equation}
and 
\begin{equation}
        \tilde{\mathbf{Y}}_{k,p} = \mathbf{L}_{k,l}\mathbf{Z}_{k,p} 0+\mathbf{L}_{k,u}\mathbf{Z}_{k,p}\mathbf{I}_{k}+0\mathbf{Z}_{k,p}.
\end{equation}

\begin{figure}[h]
    \centering
    \includegraphics[trim={0 20 0 25}, clip,width = \linewidth]{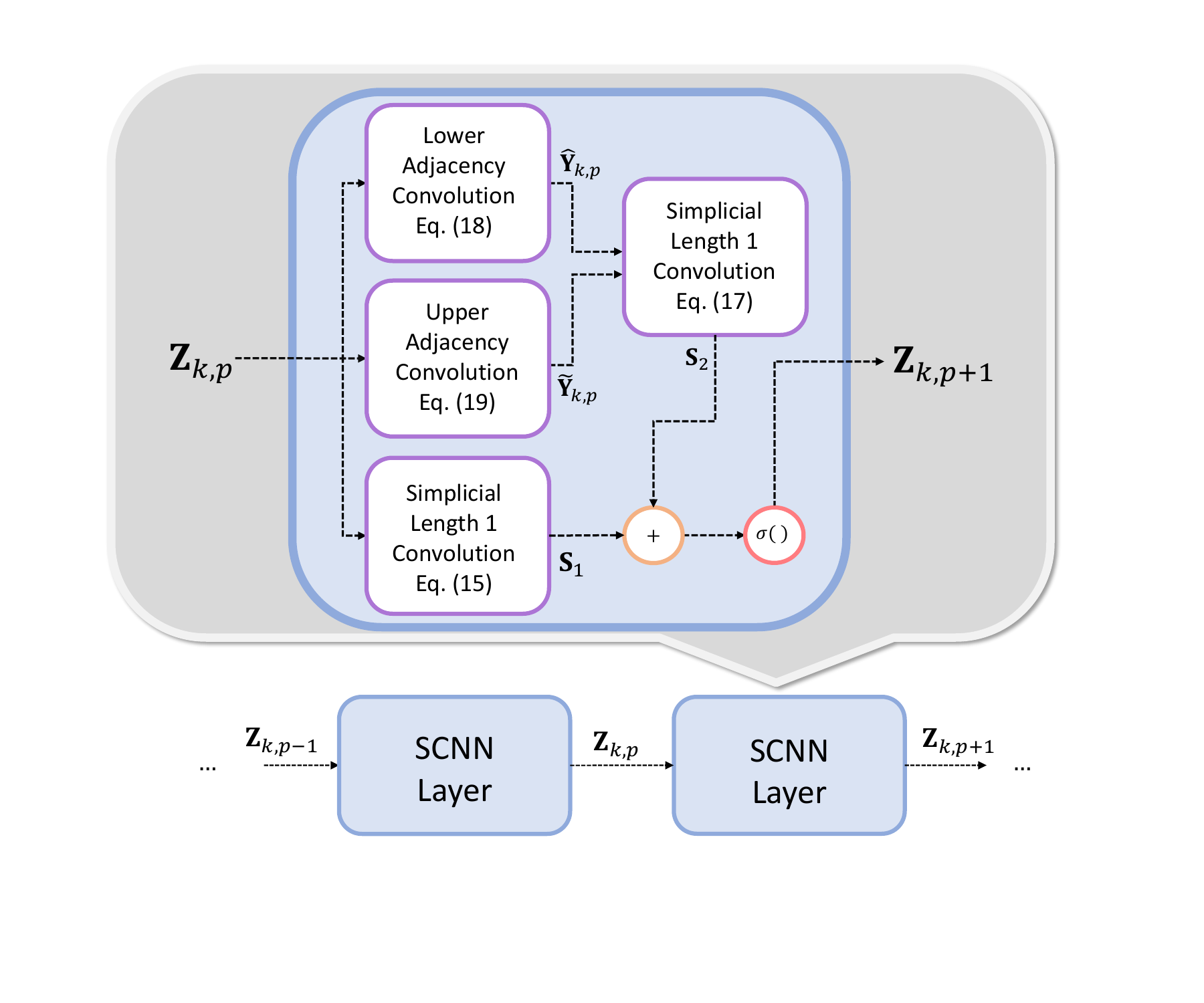}
    \caption{Decomposing a SCNN layer with $J=2$ in equation~\eqref{eq_break} into Simplicial Convolutions with $J = 1$.}
    \label{fig_break}
\end{figure}

From the above equations, it is not difficult to check that for one SCNN layer with $J=2$, it is equivalent to having a combination of $4$ SCNN layers with $J = 1$ stacked in two layers. 
The reason behind this is that when $J>1$, for each $j$ that is greater than $1$, the simplicial length $J$ convolution in \eqref{eq_break} will introduce a calculation of either $\mathbf{L}^{j}_{k, l}$ or $\mathbf{L}^{j}_{k, u}$. 
From the matrix multiplication perspective, using the lower adjacency aggregation $\mathbf{L}^{j}_{k, l}\mathbf{Z}_{k,p} = \mathbf{L}^{j-1}_{k, l}(\mathbf{L}_{k, l})\mathbf{Z}_{k,p}$ as an example, we can see that for $J = 2$, it requires \eqref{eq_break} to do the simplicial length 1 convolution twice to calculate the results. 
A similar argument for obtaining the results for upper adjacency can be made.
As a result, setting $J = 2$ in \eqref{eq_break} is essentially doing the computation of $4$ \eqref{eq_break} with $J = 1$, which introduces high computational complexity and over-smoothing when the length of convolution $J$ or the number of layers $P$ is high.
A visualization of the above breaking down of \eqref{eq_break} is provided in Figure~\ref{fig_break} for the case of $J = 2$.
Intuitively, for an SNN with a generic length $J$ that is large, a length $J$ SCNN layer can be decomposed into many length 1 SCNN layers using the above analysis.
In other words, when $J$ is large, one layer of SCNN intuitively incorporates many length 1 SCNNs, which leads to the high model and computational complexity of SCNN.

In addition to the complexity issue pointed out above, the SCNN is also prone to over-smoothing. 
To see this, let us revisit the SCNN in \eqref{eq_SCNN_multi}.
Each multiplication of the Hodge Laplacian in SCNN is a 1-hop neighbor aggregation because, by definition, the non-diagonal elements of the Hodge Laplacian matrix are non-zero only when there is a neighbor in between $v_i$ and $v_j$ two elements $i$ and $j$.  
So, multiplying the Hodge Laplacian $\mathbf{L}_k$ with a feature vector $\boldsymbol{x}_k$ essentially aggregates all the neighborhood features that are 1-hop away, and doing it for $\mathbf{L}^j_k$ will aggregate from $j$-hops away. 
For the SCNN layer in \eqref{SCNN_layer}, if $J$ is large, one single SCNN layer is possible to aggregate the entire simplex as each $j = 1...J$ will aggregate from neighbors further and further, which makes SCNN prone to over-smoothing of the features.
The over-smoothing problem is the exact analogy of what was seen in the GCNs \citep{liu2020_over_smooth} where the over-smoothed output has similar values among different elements in a simplex and causes the output to be indistinguishable.

Before we proceed with the Bi-SCNN, let us briefly discuss the Simplicial MPNN.
In short, the Simplicial MPNN increased the representation ability of SCNN by individually defining the aggregation of simplex using the concept of message. 
There are 4 types of Message Passing on a simplicial complex: the upper adjacency, the lower adjacency, the boundary adjacency, and the coboundary adjacency \citep{bodnar_2021_weisfeiler}.
In Simplicial MPNN, each of the adjacencies can be modeled as trainable weights, and the weights can be learned and updated through backpropagation.
As so, the number of trainable weights in a Simplicial MPNN layer is proportional to the number of simplices in the particular simplicial order of the adjacency of choice.
The significant increase in the number allows the Simplicial MPNN to have better representation power compared to GCN, SNN, and SCNN when properly trained.
The SNN in \eqref{eq_SNN_multi} and SCNN in \eqref{eq_SCNN_multi} can be obtained by specifying MPNN using upper and lower adjacencies.
Here is a realization of the SCNN using Simplicial MPNN:
\begin{equation}
    \mathbf{{Z}}_{k, {p+1}} = \sigma \left( \mathbf{B}_k^T\text{diag}(\boldsymbol{\gamma}_{k, p})\mathbf{B}_k\mathbf{{Z}}_{k, {p}}+\mathbf{B}_{k+1}\text{diag}(\boldsymbol{\theta}_{k, p})\mathbf{B}_{k+1}^T\mathbf{{Z}}_{k, {p}}\right),
        \label{MPPNN_multi_bi}
\end{equation}
where $\boldsymbol{\gamma}_{k, p}$ and $\boldsymbol{\theta}_{k, p}$ are the trainable parameters
Even though MPNN has high representation power, it comes at a cost of significantly increased parameter size, causing the runtime and computational complexity for training to be high. 
This also means that the Simplicial MPNN shares the same drawbacks as SNN and SCNN we pointed out earlier.

\subsection{Simplicial Feature Binarization}
Let us now examine how to simultaneously reduce the model complexity, enhance the time efficiency, and maintain the effectiveness of its simplicial embedding of the simplicial-convolution-based architectures. 
We will address each of the identified challenges systematically.

First, the complexity of the SCNN causes slow computation and over-smoothing. 
In the past, one simple yet effective solution to the over-smoothing problem of graph neural network algorithms was to reduce the number of layers. 
In order to avoid the complexity of SCNN, the proposed Bi-SCNN will be formed using purely length 1 simplicial convolution. 
In Bi-SCNN, we set the length of the filter to $J = 1$, and treat the number of layers as a hyperparameter during training to reduce over-smoothing.
It has been demonstrated on GNNS with node-based graph data that similar reductions in computation can enhance the scalability of neural networks and help mitigate over-smoothing \citep{kipf2016semi}.

To reduce the model complexity, other than simply reducing the number of parameters, one of the many solutions is Binarization.
One example of binarization is the Bi-GCN, where both the trainable weights and the features are binarized; the forward propagation is done by binary operations bit count and XOR to combine the binarized weights and features into a new node embedding. 
Aside from the fact that these node-based algorithms are not suitable to be deployed on simplicial complexes, there are a few additional problems to be addressed.
The Bi-GCN binarizes both the weights and features, leading to reduced performance under certain circumstances. 
Also, even though Bi-GCN uses binary operations to optimize the computation cost, the forward propagation is indifferent and mathematically identical compared with the original GCN. 
Another approach is seen in the Graph-Sign algorithms \citep{yan_2022_sign}, which delivers a sign-error update at each iteration that has a low run time and a low complexity. 
The drawback of the Graph-Sign algorithm is that it operates on a predefined bandlimited filter, which may require prior expert knowledge to obtain. 

In Bi-SCNN, we propose a new architecture where we combine feature normalization and feature binarization together with simplicial filters; the nonlinearity is achieved via feature normalization and feature binarization, allowing forward and backward propagation to be efficient and accurate. 
Let us revisit the multi-variate simplicial convolution in \eqref{eq_SCNN_multi}. 
Take an approach that resembles what was seen in the Bi-GCN \citep{Wang2021_Bi_GCN} by following the graph-based feature binarization step in Bi-GCN, we define the feature binarization of simplicial data $\mathbf{X}_k$ on $k$-simplex $\mathcal{S}_k$ as
\begin{equation}
    \mathbf{\bar{X}_k} \approx \mathbf{M}_k\circ \text{Sign}(\mathbf{X}_k), 
    \label{ed_binarized_feature}
\end{equation}
where $\circ$ is the Hadamard product, Sign$()$ is the Sign function and $\mathbf{\bar{X}_k}$ is the (weighted) binarized simplicial features.
As a reference, the Sign() function is defined as 
\begin{equation}
     \text{Sign(x)}= \begin{cases}    1,& \text{if } x\geq 0\\    -1,              & \text{otherwise}.
\end{cases}
\end{equation}
The matrix $\mathbf{M}_k$ acts as the weights for the binarized features and is formed by a feature normalization:
\begin{equation}
     {m_i} = \frac{norm_{1}(\mathbf{r}_k)}{d_k},
     \label{feature_norm}
\end{equation}
where $norm_{1}(\mathbf{r}_k)$ is the $l_1$-norm done on the $i^{th}$ row of $\mathbf{X}_k$ ($\mathbf{r}_k$ in \eqref{feature_norm}) and $d_k$ is the number of features in $\mathcal{S}_k$. 
All elements of the same row in $\mathbf{M}_k$ have the same value, but this is only for mathematical representation convenience. 
In practice, $\mathbf{M}_k$ is stored as a vector $\boldsymbol{m}_k$ and is broadcasted using element-wise multiplications when used to multiply with $\text{Sign}(\mathbf{X}_k)$. 
Notice that the Sign() function only outputs binary values $-1$ and $+1$ assuming that we ignore the zeros, this means that the size of the binarized features $\mathbf{\bar{X}_k}$ is reduced in practice when compared to $\mathbf{X}_k$  \citep{Wang2021_Bi_GCN}.

Now, plugging the feature binarization \eqref{ed_binarized_feature} into \eqref{eq_SCNN_multi} and setting $J = 1$, we will get 
\begin{equation}
        \mathbf{\bar{Z}}_{k, {p+1}} = \sigma \left( \mathbf{L}_{k,l} \mathbf{\bar{Z}}_{k, {p}} \mathbf{\Gamma}_{k, {p}} + \mathbf{L}_{k,u}\mathbf{\bar{Z}}_{k, {p}} \mathbf{\Theta}_{k, {p}} + \mathbf{\bar{Z}}_{k, {p}} \mathbf{\Xi}_{k, {p}}\right).
        \label{SCNN_multi_bi}
\end{equation}
Looking at the expression in \eqref{SCNN_multi_bi}, aside from benefiting from the model size reduction, \eqref{SCNN_multi_bi} still follows the exact same forward propagation strategy as SCNN.
Under unfavorable scenarios, \eqref{SCNN_multi_bi} could potentially only offer a marginal speed advantage, resulting in approximately equivalent execution times.
We would like to improve the run speed of this Binary setup further by propagating only the Sign() part and further reducing the computation.
The motivation behind this is that the Sign only has 3 outputs, with the output 0 rarely appearing in practice, so propagating only 1s and -1s significantly reduces the computation. 
Using the two properties of Hadamard products $\mathbf{A}\circ\mathbf{B} = \mathbf{B}\circ\mathbf{A}$ and $\mathbf{A}\circ\mathbf{(B+C)} = \mathbf{A}\circ\mathbf{B}+\mathbf{A}\circ\mathbf{C}$, with some algebraic derivations, equation \eqref{SCNN_multi_bi} becomes
\begin{equation}
\begin{split}
    \mathbf{\bar{Z}}_{k, {p+1}} = & \mathbf{M}_{k, p}\circ ( \mathbf{L}_{k,l} \text{Sign}(\mathbf{\bar{Z}}_{k, {p}}) \mathbf{\Gamma}_{k, {p}} + \\
    & \mathbf{L}_{k,u} \text{Sign}(\mathbf{\bar{Z}}_{k, {p}}) \mathbf{\Theta}_{k, {p}} + \text{Sign}(\mathbf{\bar{Z}}_{k, {p}}) \mathbf{\Xi}_{k, {p}} ). 
        \label{SCNN_multi_bi_3}
\end{split}
\end{equation}

Notice that the $l_1$-norm is nonnegative, so at each layer, $\mathbf{M}_{k, p}$ can be viewed as $\mathbf{M}_{k, p} =$ ReLu($\mathbf{M}_{k, p}$). 
The input of the $p+1^{th}$ layer in \eqref{SCNN_multi_bi_3} will binarize the output of the $p^{th}$ layer regardless of the magnitude, so we can safely input only the binarized simplicial features (without the weights $\mathbf{M}_k$) to the $p+1^{th}$ layer when $p>1$. 
Also, we should point out that the Sign() function is non-linear. 
In this way, the nonlinearity in both the Sign() function and the $l_1$-norm can serve as activations in forward propagation. 
Additionally, both paths will benefit from a reduction in computational complexity, which we will be analyzing in the next subsection.

\begin{figure}[t]
\centering
\includegraphics[trim={0 20 0 25}, clip, width=\linewidth]{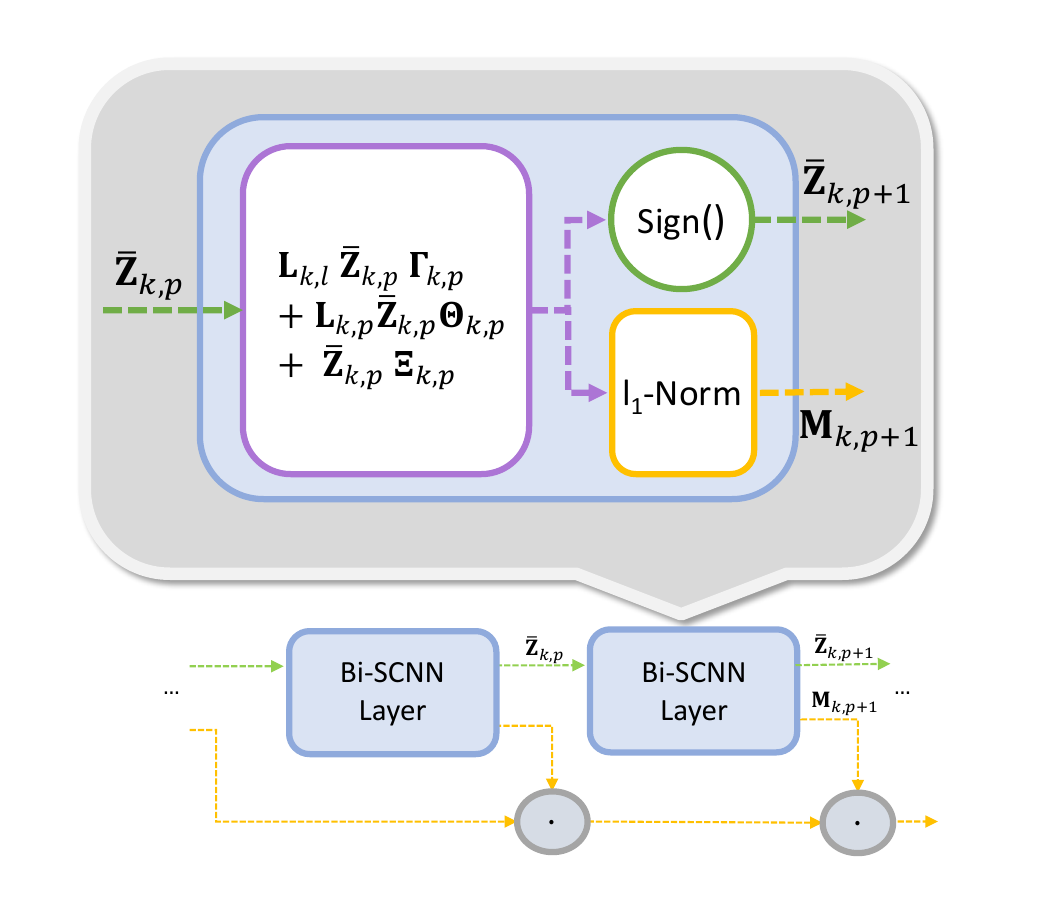}
\caption{An illustration of one Bi-SCNN layer.}
\label{Fig_Bi_SCNN_layer}
\end{figure}

\begin{figure*}[htb]
\centering
\includegraphics[trim={0 60 0 60},width=\textwidth]{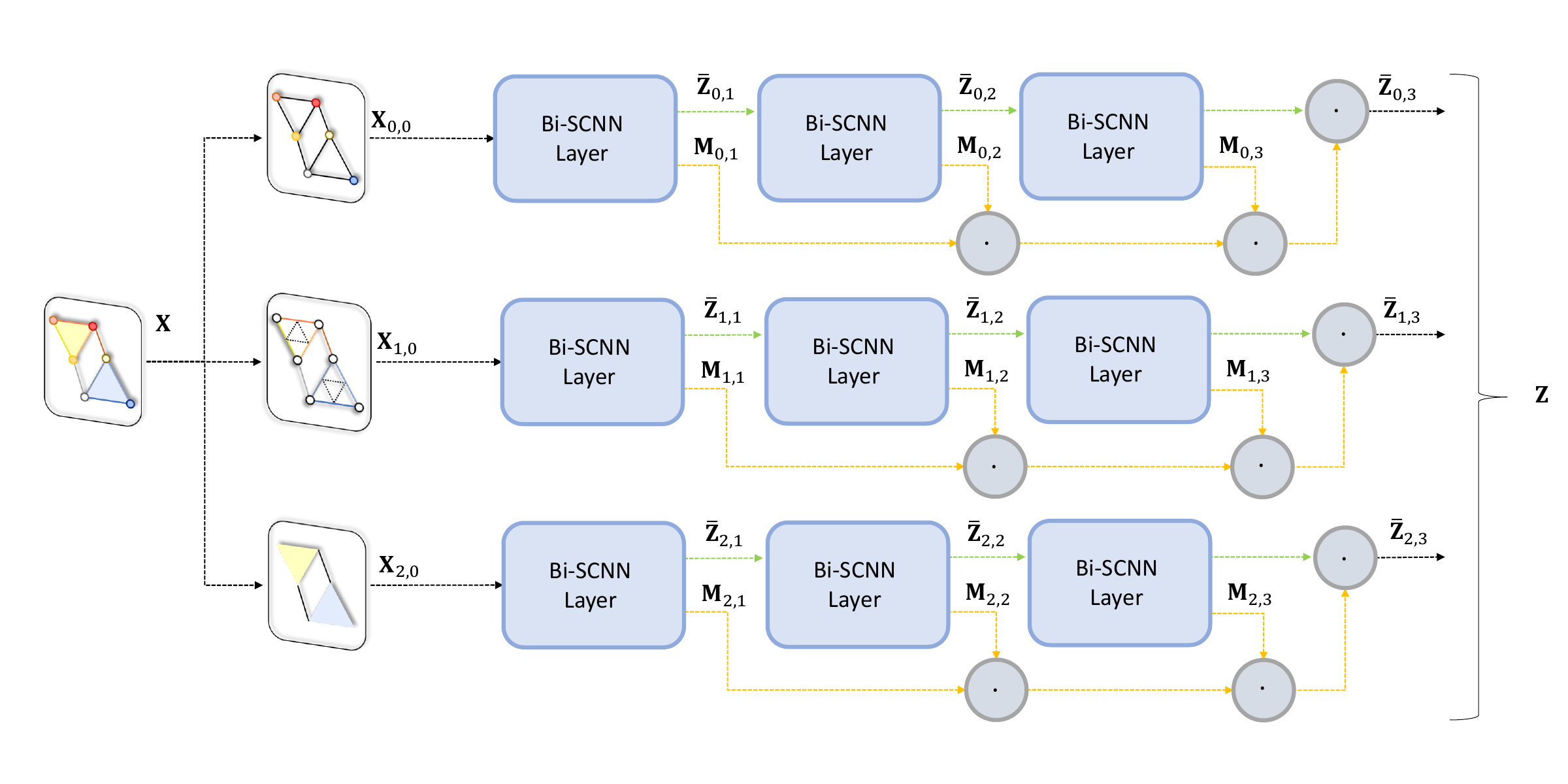}
\caption{An illustration of a $P$-layer Bi-SCNN network in \eqref{Bi_SCNN_layer_2} for simplicial order $K = 2$ and $P = 3$. The forward propagation is indicated in the dashed arrows with each colored path corresponding to the component of the same color in Figure~\ref{Fig_Bi_SCNN_layer}. For the input $\mathbf{X}$, the adjacencies within a simplicial signal $\mathbf{X}_k$ are denoted in black.}
\label{Fig_Bi_SCNN_network}
\end{figure*}

The forward propagation of a series of $P$ concatenated Bi-SCNN is obtained by rearranging \eqref{SCNN_multi_bi_3} a using the property $\mathbf{A}\circ\mathbf{B} = \mathbf{B}\circ\mathbf{A}$ again: 
\begin{equation}
    \mathbf{\bar{Z}}_{k, {P}} = \mathbf{M}_{k, P}\circ\mathbf{M}_{k, P-1} \circ ... \circ \mathbf{M}_{k, 2} \circ \mathbf{Q}_{k, P-1}, 
        \label{Bi_SCNN_layer_2}
 \end{equation}
where for the $p^{th}$ layer 
\begin{equation}
    \mathbf{Q}_{k, p} = \text{Sign}(\mathbf{L}_{k,l} \mathbf{\bar{Z}}_{k, p} \mathbf{\Gamma}_{k, {p}} + 
    \mathbf{L}_{k,u}\mathbf{\bar{Z}}_{k, p} \mathbf{\Theta}_{k} + \mathbf{\bar{Z}}_{k, p} \mathbf{\Xi}_{k}).
\end{equation}
The way to achieve the forward propagation of Bi-SCNN in \eqref{Bi_SCNN_layer_2} in practice is that each Bi-SCNN layer will have two outputs: the feature normalization $\mathbf{M}_{k, p}$ and feature binarization $\mathbf{Q}_{k, p}$.
Within the two outputs, notice that only $\mathbf{Q}_{k, p}$ will be forward-propagated to the next Bi-SCNN layer, meaning that $\mathbf{Q}_{k, p} = \mathbf{\bar{Z}}_{k,p}$, with the following two exceptions: 1. at the first layer where $\mathbf{X}_k$ is the input, and 2. the final layer all the $\mathbf{M}_{k, p}$ are multiplied together at the end to form $\mathbf{\bar{Z}}_{k, {P}}$. 
Each of the normalization $\mathbf{M}_{k, p}$ is propagated straight to the final Bi-SCNN layer and is combined using element-wise multiplication. 
An illustration of one Bi-SCNN layer is shown in Fig.\ref{Fig_Bi_SCNN_layer}, and an illustration of a Bi-SCNN network with  $P=3$ is shown in Fig.\ref{Fig_Bi_SCNN_network}.

\subsection{Interpretation and Analysis of Bi-SCNN }

Intuitively, a single Bi-SCNN layer on a $k$-simplex can be understood as using the (upper and lower) Hodge Laplacians to aggregate the weighted simplicial features with the addition of the harmonic component.
From the spatial TSP perspective, the weights are trained to optimize a binary-sign feature aggregation with a matrix that records the magnitude of each aggregation. 
From the spectral TSP perspective, the goal of training a Bi-SCNN is to obtain a filter for the binary-sign update that best fits the spectrum of the training data. 

Aside from the fact that all simplicial-convolution-based algorithms benefit from the sparsity of $\mathbf{L}_k$, let us look into \eqref{Bi_SCNN_layer_2} to see the high time-efficiency of Bi-SCNN. 
In practice, Sign() is approximated by a hard tanh function, meaning that a percentage $r_k$ of the matrix $\mathbf{\bar{Z}}_{k, {p}}$ consists only of $1$s and $-1$s.
In this case, $r_k$ element-wise multiplications can be omitted in the matrix multiplication, reducing $N_k^2 d_k$ multiplications to $(1-r_k) N_k^2 d_k$ multiplications. 
Assuming each multiplication is 1 FLOP, each Bi-SCNN layer is at least $r_k$ FLOPS faster than the SCNN layer for the $2^{nd}$ layer and beyond. 

Even though the Sign() used in the Bi-SCNN layers \eqref{Bi_SCNN_layer_2} has no derivative at 0 due to its discontinuity at $0$ and a derivative of $0$ for all other values, we approximate Sign() using a hard tanh function, allowing us to conduct forward propagation and backward propagation as usual. 
This approach is known as the straight-through estimator, where gradient-based optimization is enabled on non-differentiable functions by using a proxy gradient during backward propagation \citep{yin2019understanding,le2022adaste}. 

Let us compare the model complexity of Bi-SCNN with its predecessors. In the actual implementation of SCNN, to compute $\mathbf{Z}_k^{p}$, $\mathbf{L}_{k,l}$, $\mathbf{L}_{k,u}$, and $\mathbf{I}_k$ are concatenated into a $3N_k \times N_k$ matrix for each $j = 0...J$; then $J$ matrices are stacked into a $d_p \times N_k \times J$ tensor. 
Trainable weights $\mathbf{\Theta}_k$, $\Gamma_{k,j}$, and $\Xi_{k}$ are stacked into a $d_{k,p} \times d_{k,p+1} \times J$ tensor in a similar manner.
Afterward, in SCNN, $\mathbf{Z}_{k,p+1}$ is computed by the Einstein summation between two tensors (aggregation and weight tensors) of size $d_p \times N_k \times J$ and $d_{k,p} \times d_{k,p+1} \times J$ back into a matrix of size $N_k \times d_{p+1}$. 
This matrix concatenation, stacking, and Einstein summation approach of the SCNN is less time efficient than direct matrix multiplication in Bi-SCNN considering we are using binarized features, especially when $J>1$. 
The SNN in \eqref{eq_SNN_multi} has a similar model complexity to the SCNN in \eqref{eq_SCNN_multi} because instead of splitting the Hodge Laplacian matrix $\mathbf{L}_k$ using Hodge decomposition, SNN kept the Hodge Laplacian matrix as a whole. 
However, it still has a length $J$ component in each SNN layer. 
Compared to SCNN and Bi-SCNN, the SNN has lower representation power because SNN does not adopt the Hodge decomposition in \eqref{eq_hodge_decomposition} to model upper, lower, and harmonic components separately. Additionally, SNN has high computation and memory demands similar to the SCNN. 

On the other hand, we should emphasize again that $\mathbf{M}_{k, p}$ is implemented using a vector in practice, followed by element-wise multiplication instead of matrix multiplication, making the Bi-SCNN more time-efficient. 
Lastly, we have mentioned earlier that the Sign() function is being approximated by a hard tanh function, which is a straight-through estimator, allowing the backward propagation calculation to be extremely efficient as straight-through estimators rely only on low-computational complexity or predefined values for the gradient calculation.

\section{Experiment Results and Discussion}
\label{sec_experiment}
\begin{figure}[h]
\centering
\includegraphics[trim={2cm 2cm 2cm 2cm},clip,width=0.9\linewidth]{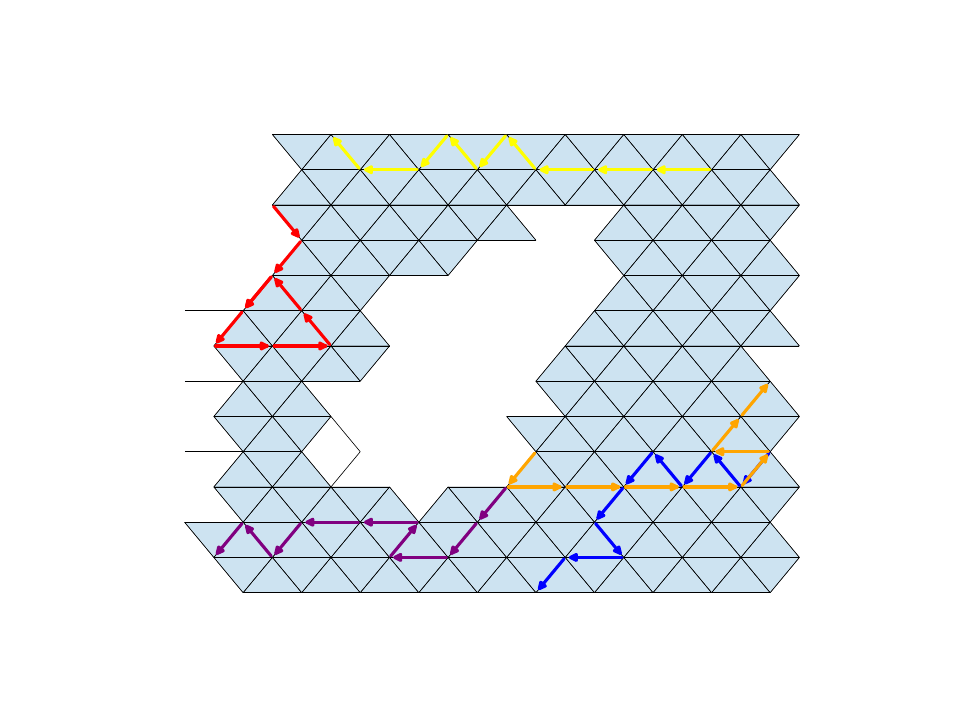}
\caption{The ocean drifter complex with illustrations of 5 different ocean drifter trajectories.}
\label{drifters}
\end{figure}

The Citation Complex \citep{ebli2020_SNN} and the Ocean Drifter Complex \citep{Schaub_2020_Random} are chosen because both datasets are already preprocessed to have features defined in simplexes with order $k\geq1$, making them suitable choices to test the performance of an algorithm on simplicial complexes. 
We would like to verify that the Bi-SCNN has a faster execution time while maintaining good accuracy when compared against existing simplicial algorithms. 
The experiments are conducted on one Nvidia RTX 3090 GPU with PyTorch version 1.8.0 and CUDA version 11.1. 
None of the two datasets provided a validation set so the experiments are tuned based on the training performance. 
All the experiments were repeated 10 times; we calculated the mean and standard derivation of the metrics. 
A visualization of the ocean drifter complex with illustrations of 5 different ocean drifter trajectories is shown in Figure~\ref{drifters}.


        
\begin{figure*}[htbp]
    \centering
    \includegraphics[trim={0 280 0 270},clip,width=\linewidth]{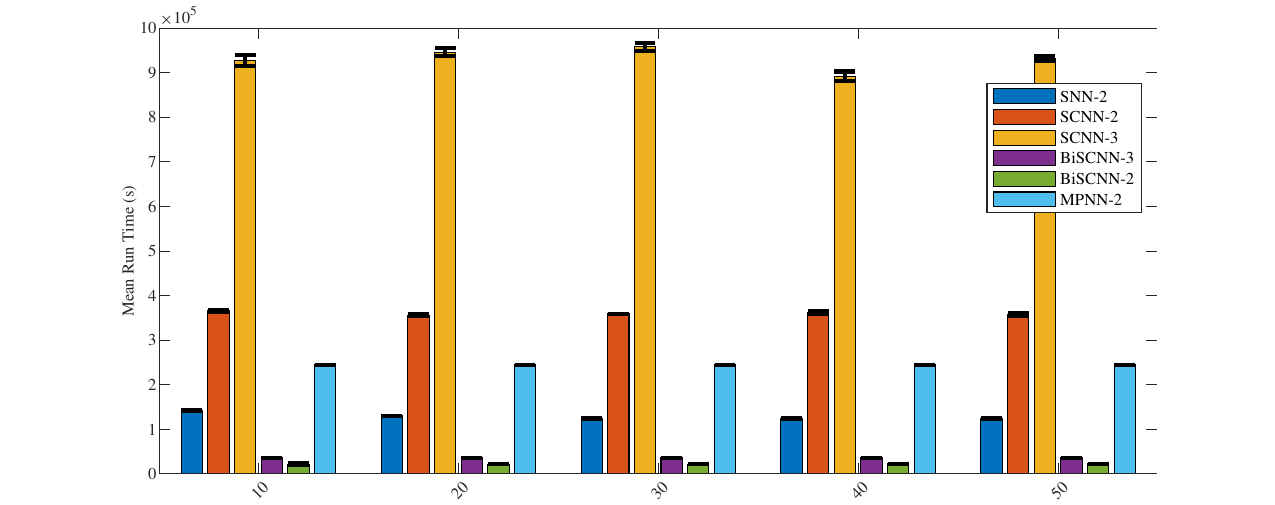}
    \caption{Mean run time of each algorithm on the Citation Complex for $10\%-50\%$ missing data.}
    \label{fig_time}
\end{figure*}

\begin{table*}[htbp]
    \centering
    \begin{tabular}{c c c c c c c} 
    \toprule
    & SNN-2 & SCNN-2 & SCNN-3 &  Bi-SCNN-3 & Bi-SCNN-2 & MPNN-2\\
        \midrule
        \multicolumn{7}{c}{10$\%$ missing}\\
            $\mathcal{S}_{0}$ &90.60$\pm$3.47 & 90.51$\pm$0.34 & 90.57$\pm$0.38& 84.52$\pm$17.95 & \bf{90.65$\pm$0.32} & 6.13$\pm$3.07\\
            $\mathcal{S}_{1}$ &91.00$\pm$02.45 & 91.02$\pm$0.25 & 90.52$\pm$1.43 & 90.77$\pm$0.24 & \bf{91.03$\pm$0.24} & 72.84$\pm$36.42\\
            $\mathcal{S}_{2}$ &90.94$\pm$05.59 & 90.60$\pm$0.67 & \bf{91.22$\pm$0.16} & 87.81$\pm$01.91 & \bf{91.22$\pm$0.16}&36.50$\pm$44.70\\
            $\mathcal{S}_{3}$ &91.45$\pm$3.828 & 91.37$\pm$0.14 & \bf{91.58$\pm$0.12} & 89.74$\pm$1.65 & \bf{91.58$\pm$0.12} & 38.17 $\pm$ 43.92\\ 
            $\mathcal{S}_{4}$ &87.89$\pm$11.59 & 91.65$\pm$0.56 & 91.88$\pm$0.23 & 85.04$\pm$16.80  & \bf{91.92$\pm$0.20} & 38.16 $\pm$ 40.48\\
            $\mathcal{S}_{5}$ &83.61$\pm$18.65 & 92.07$\pm$0.33 & 82.96$\pm$27.64 & 89.44$\pm$0.49  & \bf{92.21$\pm$0.23} & 22.17 $\pm$ 0.00\\
        \midrule 
        \multicolumn{7}{c}{20$\%$ missing}\\
            $\mathcal{S}_{0}$ &81.22$\pm$0.63 & 81.36$\pm$0.64 & 81.14$\pm$0.63  & 81.08$\pm$0.61 & \bf{81.39$\pm$0.66} & 3.07 $\pm$ 3.76\\
            $\mathcal{S}_{1}$ &82.17$\pm$0.32 & 82.16$\pm$0.32 & 69.76$\pm$26.24 & 82.07$\pm$0.25 & \bf{82.20$\pm$0.31} & 49.38 $\pm$ 40.14\\
            $\mathcal{S}_{2}$ &82.42$\pm$0.49 & 82.26$\pm$0.51 & 74.42$\pm$24.81 & 80.53$\pm$0.58 & \bf{82.68$\pm$0.27} & 62.33 $\pm$ 33.15\\
            $\mathcal{S}_{3}$ &83.11$\pm$0.38 & 82.97$\pm$0.26 & \bf{83.23$\pm$0.15} & 81.81$\pm$1.28 & \bf{83.23$\pm$0.15} & 33.27 $\pm$ 40.74\\ 
            $\mathcal{S}_{4}$ &75.63$\pm$23.65 & 83.47$\pm$0.40 & 81.84$\pm$5.39& 82.79$\pm$0.45 & \bf{83.64$\pm$0.19}&18.33 $\pm$ 32.92\\
            $\mathcal{S}_{5}$ &73.03$\pm$25.38 & 84.22$\pm$0.27 & 75.88$\pm$25.30 & 66.45$\pm$2.69 & \bf{84.34$\pm$0.23}&22.17 $\pm$ 0.00\\
        \midrule 
        \multicolumn{7}{c}{30$\%$ missing}\\
            $\mathcal{S}_{0}$ &72.16$\pm$0.66 & 72.27$\pm$0.57 & 71.82$\pm$1.28 & 72.05$\pm$0.61& \bf{72.33$\pm$0.50} & 6.90 $\pm$ 2.30\\
            $\mathcal{S}_{1}$ & 73.15$\pm$0.42 & 73.13$\pm$0.44 & 72.87$\pm$0.52 & 72.64$\pm$0.79 & \bf{73.98$\pm$0.41} & 21.91 $\pm$ 33.44\\
            $\mathcal{S}_{2}$ &73.73$\pm$0.67 & 73.46$\pm$0.51 & 66.70$\pm$21.96 & 72.20$\pm$0.32 & \bf{73.98$\pm$0.26} & 49.79 $\pm$ 33.08\\
            $\mathcal{S}_{3}$ &74.70$\pm$0.30 & 74.52$\pm$0.21 & 67.29$\pm$22.43 & 72.44$\pm$0.40 & \bf{74.78$\pm$0.16} & 37.48 $\pm$ 34.26\\ 
            $\mathcal{S}_{4}$ &68.31$\pm$21.98 & 75.56$\pm$0.26 & 68.05$\pm$22.69 & 74.73$\pm$08.27 & \bf{75.68$\pm$0.20} & 30.47 $\pm$ 36.92\\
            $\mathcal{S}_{5}$ &72.09$\pm$10.87 & 76.52$\pm$0.31 & \bf{76.63$\pm$0.24} & 65.00$\pm$18.89 & \bf{76.63$\pm$0.23} & 22.17 $\pm$ 0.00\\
        \midrule 
        \multicolumn{7}{c}{40$\%$ missing}\\
            $\mathcal{S}_{0}$ & 62.36$\pm$0.72 & 62.78$\pm$0.65 & 62.07$\pm$1.83 & 62.59$\pm$0.61 & \bf{62.81$\pm$0.65} & 3.84 $\pm$ 3.84\\
            $\mathcal{S}_{1}$ & 64.10$\pm$0.28 & \bf{64.10$\pm$0.28} & 63.74$\pm$0.91 & 63.57$\pm$0.79 & \bf{64.10$\pm$0.28}& 44.93 $\pm$ 29.42\\
            $\mathcal{S}_{2}$ & 64.73$\pm$0.79 & 64.54$\pm$0.63 & \bf{65.52$\pm$0.20} & 63.86$\pm$0.32 & 65.17$\pm$0.20 & 45.62 $\pm$ 29.86\\
            $\mathcal{S}_{3}$ & 66.26$\pm$0.28 & 66.07$\pm$0.33 & \bf{66.31}$\pm$0.21 & 65.40$\pm$0.40 & \bf{66.31}$\pm$0.21 & 19.91 $\pm$ 30.41\\ 
            $\mathcal{S}_{4}$ & 66.89$\pm$1.66 & 67.33$\pm$0.39 & 50.26$\pm$27.14 & 66.85$\pm$0.83 & \bf{67.57$\pm$0.23} & 29.42 $\pm$ 31.82\\
            $\mathcal{S}_{5}$ & 61.98$\pm$14.03 & 68.86$\pm$0.18 & 62.21$\pm$20.31 & 53.91$\pm$18.89 & \bf{68.98$\pm$0.18} & 22.17 $\pm$ 0.00\\
        \midrule 
        \multicolumn{7}{c}{50$\%$ missing}\\
            $\mathcal{S}_{0}$ & 54.01$\pm$0.54 & 54.03$\pm$0.06 & 53.98$\pm$0.58 & 50.28$\pm$5.35 & \bf{54.18$\pm$0.52} & 2.30 $\pm$ 3.52\\
            $\mathcal{S}_{1}$ & 55.05$\pm$0.39 & 55.08$\pm$0.04 & 54.33$\pm$1.37 & 54.21$\pm$0.94 & \bf{55.81$\pm$0.40} & 37.99 $\pm$ 24.95\\
            $\mathcal{S}_{2}$ & 55.32$\pm$2.13 & 55.78$\pm$0.06 & \bf{56.27}$\pm$0.24 & 52.95$\pm$4.04 & \bf{56.27}$\pm$0.24 & 28.12 $\pm$ 28.12\\
            $\mathcal{S}_{3}$ & 57.81$\pm$0.31 & 57.33$\pm$0.03 & 57.82$\pm$0.28 & 57.19$\pm$0.37 & \bf{57.85$\pm$0.25} & 23.13 $\pm$ 28.33\\ 
            $\mathcal{S}_{4}$ & 57.64$\pm$4.43 & 59.05$\pm$0.04 & \bf{59.54}$\pm$0.29 & 57.98$\pm$3.21 & \bf{59.54}$\pm$0.29 & 16.03 $\pm$ 24.94\\
            $\mathcal{S}_{5}$ & 49.87$\pm$16.51 & 60.72$\pm$0.09 & \bf{61.15}$\pm$0.19 & 55.39$\pm$6.39 & \bf{61.15}$\pm$0.19 & 22.17 $\pm$ 0.00\\
        \bottomrule

    \end{tabular}
        \caption{Experiment accuracy on citation complex from 10$\%$ to 50$\%$ data missing rate. The best-performing model (including ties) of each simplex is in bold.}
    \label{accuracy_citation}
\end{table*}

\begin{table*}[htbp]
    \centering
    \begin{tabular}{c c c c c c c} 
        \toprule
         & SNN-2 & SCNN-2 & SCNN-3 & Bi-SCNN-3 & Bi-SCNN-2 & MPNN-2\\
         \midrule
            10$\%$ missing & 141.78$\pm$0.37 s & 365.92$\pm$2.23 s&  927.38$\pm$11.58 s & 35.96$\pm$0.18  s& \bf{21.21} $\pm$0.78 s & 245.10 $\pm$ 0.21 s\\
            20$\%$ missing & 130.25$\pm$0.19 s & 356.38$\pm$1.87 s& 945.06$\pm$8.42 s & 36.23$\pm$0.20 s& \bf{22.23} $\pm$1.73 s & 245.02 $\pm$ 0.13 s\\
            30$\%$ missing & 124.06$\pm$0.61 s & 358.98$\pm$3.11 s& 957.68$\pm$9.04 s & 36.15$\pm$0.20 s& \bf{22.20} $\pm$1.35 s & 244.83 $\pm$ 0.12 s\\
            40$\%$ missing & 124.85$\pm$0.41 s & 362.22$\pm$2.78 s& 892.22$\pm$9.86 s & 35.94$\pm$0.24 s& \bf{22.46} $\pm$1.12 s & 244.87 $\pm$ 0.12 s\\
            50$\%$ missing & 124.33$\pm$0.22 s & 357.77$\pm$2.80 s& 931.48$\pm$5.71 s & 36.11$\pm$0.24 s& \bf{22.38} $\pm$1.18 s & 244.99 $\pm$ 0.15 s\\
    \bottomrule  
    \end{tabular}
        \caption{Model execution times in seconds (forward and backward propagation only, discarding pre-processing and post-processing) on citation complex with $k=0 ... 5$, from 10$\%$ to 50$\%$ data missing rate. The best-performing model is in bold.}
        \label{time_citation}
\end{table*}

\subsection{Citation Complex}
The raw data of the Citation Complex is from Semantic Scholar Open Research Corpus \citep{ammar2018construction}, where the number of citations of $k+1$ authors forms a $k$-simplex where $k = 0...5$. 
The 6 simplices naturally form a simplicial complex because they form a hierarchical structure: each $k$-simplex is a subset author collaborations of the $k+1$-simplex.  
The number of data elements in each simplicial order for $k = 0 ... 5$ is $ N_0 = 352$, $N_1 = 1474$, $N_2 = 3285$, $N_3 = 5019$, $N_4 = 5559$, and $N_5 = 4547$.
It is worth mentioning that each simplicial complex standing alone by itself could be viewed as a dataset as well.
For example $\mathcal{S}_0$ is a graph with data on the nodes only, which means that GNNs such as the GAT and the GCN can be deployed on it. 
However, graph node algorithms are only suitable for data defined on the nodes, and for higher-order structures like the data on the graph edges or triangles formed by the edges, we still need to rely on simplicial complex algorithms.

To examine the performance of our Bi-SCNN algorithm, we will follow the training and testing procedure seen in SNN  \citep{ebli2020_SNN} and SCNN \citep{Yang_2022_SCNN}. 
For each $k$-simplex a portion of the simplicial data will be missing at random. 
In this transductive learning task, the goal is to impute the missing data from the existing data.
We will be testing 5 different missing rates: 10\%, 20\%, 30\%, 40\%, and 50\%. 

A 2-layer Bi-SCNN (Bi-SCNN-2) and a 3-layer Bi-SCNN (Bi-SCNN-3) will be compared against a 2-layer SNN (SNN-2), a 2-layer SCNN (SCNN-2), a 3-layer SCNN (SCNN-3), and a 2-layer Simplicial MPNN (MPNN-2). 
The number of simplicial filters per layer is set to 30 for all algorithms except for MPNN-2. 
Each layer in SNN-2 and SCNN-2 will have a length $J = 1$ to keep the model size roughly the same order of magnitude as the Bi-SCNN-2. 
SCNN-3 will have a filter length $J = 2$, which is the best-performing setting seen in the original paper \citep{Yang_2022_SCNN}. 
GCN is not included because SNN is equivalent to the GCN for $k = 0$.
Due to the memory demand of the MPNN-2 and the hardware constraint of our experiment machine, MPNN-2 will have 2 layers but each layer only has 4 simplicial filters to avoid out-of-memory (OOM) error. 
In theory, this will not affect the MPNN by much as we can represent the combination of $J$ filters using a single filter using TSP techniques: $\sum_{j = 1}^{J}\mathbf{U}_k h^{(j)}(\mathbf{\Lambda}_k) \mathbf{U}_k^T = \mathbf{U}_k \mathbf{\Sigma}_k \mathbf{U}_k^T$, where $\mathbf{\Sigma}_k = \sum_{j = 1}^{J} h^{(j)}(\mathbf{\Lambda}_k)$.

\begin{figure}[h]
    \centering
    \includegraphics[trim={180 330 180 300},clip,width=\linewidth]{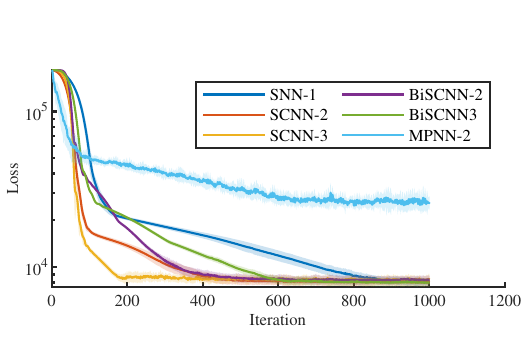}
    \caption{The mean training loss for each algorithm on the Citation Complex under $10\%$ missing rate.}
    \label{fig_loss}
\end{figure}

We follow the parameter optimization seen in SCNN \citep{Yang_2022_SCNN} where all algorithms use Adam optimizer with a learning rate of $0.001$ and minimize the $l_1$-loss on the training set for 1000 iterations. 
To make the comparison fair, all the tested algorithms use purely Simplicial Layers except for the activation functions.  
The accuracy of the imputed missing data will be considered as correct if it is within $\pm1\%$ of the ground truth value. 
The experiment run time is recorded on the total run time of all the forward and backward propagation. 
The accuracy of each algorithm setting is shown in Table~\ref{accuracy_citation}. 
The run time of the experiments is recorded in Table~\ref{time_citation}; there is also a visual illustration of the run time for 10$\%$ missing rate in Figure~\ref{fig_time}. 
A summary of the number of trainable parameters of each algorithm can be found in Table~\ref{citation_parameter_size}. 
The mean training loss of the 10 experiment runs of all the tested algorithms on the Citation Complex with $10\%$ missing data is shown in Fig.~\ref{fig_loss}.

\begin{table}[htbp]
    \centering
    \begin{tabular}{c c} 
    \toprule
    Model & Trainable Parameter \\
    \midrule
    SNN-2 & 906 \\
    SCNN-2 & 1986 \\
    SCNN-3 & 29166 \\
    Bi-SCNN-2 & 1146 \\
    Bi-SCNN-3 & 15726 \\
    MPNN-2 & 503463946 \\
    \bottomrule
    \end{tabular}
    \caption{The number of trainable parameters for each model on the Citation dataset.}  
    \label{citation_parameter_size}
\end{table}

First, analyzing the accuracy in Table~\ref{accuracy_citation}, we can see that SNN, SCNN, and Bi-SCNN have similar performances. 
This is because all three algorithms are based on simplicial convolutions. 
However, further analyzing Table~\ref{accuracy_citation}, we can see that 2-layer Bi-SCNN performs the best in most of the simplices under all the tested missing rates. 
This confirms that even though the Bi-SCNN uses a simpler propagation scheme by only having a length-1 simplicial filter, the embedding power is equivalent to the SNN or SCNN in the current experiment. 
We think this is due to the fact that the 3-layer models have too many simplicial convolutional filters and the simplicial embedding got over-smoothed during the propagation process.

The performance of the MPNN is suboptimal due to the extensive number of trainable parameters, necessitating additional preprocessing techniques or implementation tricks to effectively train the architecture. 
Analyzing the output of MPNN, we found out that for $\mathcal{S}_0$ and $\mathcal{S}_5$, the MPNN over-smoothed the output: the majority of the output shares a similar value. 
This can be again confirmed by the low standard deviation of the accuracies.
Another interesting observation is that the 2-layer models perform slightly better than the 3-layer models in many cases. 
As for $\mathcal{S}_1$, $\mathcal{S}_2$, $\mathcal{S}_3$, and $\mathcal{S}_4$, the high standard deviation can be caused by the following 3 reasons: 1. the parameter size and model complexity is too high compared to the data size; 2. the training duration is relatively too short compared to the parameter size; 3. poor initialization of parameters.

Looking at Figure~\ref{fig_loss}, we can see that all the tested algorithms converge and their convergence behaviors are distinctively different. 
The different convergence behavior is an indication that SNN, SCNN, and Bi-SCNN have different simplicial representations. 
For MPNN-2, the relatively high loss can be attributed to an excessive number of trainable parameters, which makes MPNN difficult to be properly trained for only 4 filters, leading to high training loss. 

If we look at the ratio of parameter size to run time, we can see that the Bi-SCNN is the most efficient among all the tested algorithms.
Bi-SCNN-2 has 20\% more parameters than SNN-2 but takes only $1/5$ the time that SNN needs to complete the experiment. 
We should point out that for roughly similar model sizes, the Bi-SCNN has higher accuracy in imputing the citations compared to the SNN because Bi-SCNN assigns weights separately by considering upper and lower simplicial adjacencies separately but SNN does not, making the Bi-SCNN more expressive. 
The Bi-SCNN is indeed a more concise model than the SNN and the SCNN because not only does Bi-SCNN use fewer parameters but also operates faster. 
The reason that the Bi-SCNN is time-efficient is that the binary-sign propagation takes fewer computations, which is proven in traditional CNN \citep{Courbariaux2016_Bi_NN} and has also been proven when data is purely on the nodes of the graph by the G-Sign \citep{yan_2022_sign} and Bi-GCN \citep{Wang2021_Bi_GCN}. 
We can conclude that the Bi-SCNN can effectively and efficiently impute the missing citations compared to the MPNN, SNN, and SCNN.  

\begin{table}[htbp]
    \centering
    \begin{tabular}{ccccc} 
    \toprule
         & Id & LR & Tanh\\
    \midrule
    SAT-2 & 60.25$\pm$6.47 & 61.75$\pm$7.24 & 64.00$\pm$6.04 \\
    
    SAT-3 & 64.75$\pm$5.30 & 62.25$\pm$2.08 & 66.50$\pm$5.50 \\
    
    SCNN-2 & 64.00$\pm$3.39 & 77.00$\pm$4.00 & 74.75$\pm$6.75 \\
    
    SCNN-3 & 69.00$\pm$2.78 & \bf{81.75$\pm$2.97} & \underline{75.25$\pm$6.66} \\
    
    Bi-SCNN-2 & \bf{71.25$\pm$4.00} & \underline{79.50$\pm$3.67} & \bf{78.75$\pm$5.94} \\
    
    Bi-SCNN-3 & \underline{71.00$\pm$4.06} & 77.00$\pm$4.15 & 73.75$\pm$3.01 \\
    \midrule
    MPNN* & 46.5$\pm$5.7& 73.0$\pm$2.7& 72.5$\pm$0.0\\
    \bottomrule
    \end{tabular}
    \caption{Classification (test) accuracy on the Ocean Drifter Dataset. The number after the activation denotes how many simplicial layers. (Bold denotes best-performing; underline denotes second-to-the-best. MPNN results are obtained from \citep{bodnar_2021_weisfeiler}).}
    \label{ocean_accuracy}
\end{table}



\begin{table*}[h]
    \centering
    \begin{tabular}{cccc}
    \toprule
         & Id & LR & Tanh\\
    \midrule
    SAT-2 & 1196.90 $\pm$ 114.80 s & 1116.93 $\pm$ 51.48 s & 1161.90 $\pm$ 28.95 s \\
    SAT-3 & 1515.73 $\pm$ 5.97 s & 1385.93 $\pm$ 1.77 s & 1491.28 $\pm$ 11.58 s \\
    SCNN-2 & 2197.16 $\pm$ 8.93 s & 2187.98 $\pm$ 1.12 s & 2196.65 $\pm$ 2.07 s \\
    SCNN-3 & 6451.00 $\pm$ 6.47 s & 6441.42 $\pm$ 10.62 s & 6399.48 $\pm$ 2.83 s \\
    Bi-SCNN-2 & \bf{149.33 $\pm$ 0.87 s} & \bf{151.06 $\pm$ 0.28 s} & \bf{150.21 $\pm$ 0.50 s} \\
    Bi-SCNN-3 & \underline{231.91 $\pm$ 0.84 s} & \underline{232.54 $\pm$ 0.46 s} & \underline{234.99 $\pm$ 0.19 s} \\
    MPNN-2 (ours) & 1646.23 $\pm$ 0.30 s& 1653.04 $\pm$ 2.08 s & 1644.69 $\pm$ 3.12 s\\
    \bottomrule
    \end{tabular}
    \caption{Experiment run time (in seconds) on the Ocean Drifter Dataset. }
    \label{ocean_time}
\end{table*}

\subsection{Ocean Drifter Complex}
The Ocean Drifter Complex dataset utilizes simplicial embedding to represent real-world ocean drifters around Madagascar Island \citep{Schaub_2020_Random}.
The ocean surrounding Madagascar Island is segmented into hexagonal grids, with each grid represented as a node in the $0$-simplex. 
If the trajectory of an ocean drifter passes two adjacent hexagonal grids an edge will be formed between the node. 
In the dataset, there are two types of trajectories and the goal is to classify which type of trajectory the input trajectory belongs to. 
The simplicial complex of the dataset contains $N_0 = 133$ nodes, $N_1 = 320$ edges, and $N_2 = 186$ triangles; the simplicial complex is shared among all the trajectories.
A visualization of this dataset is shown in Figure~\ref{drifters}. 
We should emphasize that the formed simplex structure itself is undirected, but the data that resides on the simplex has orientations. 
The data provided 160 training trajectories and 40 testing trajectories all on the $1$-simplex. 
In other words, the feature only exists on the edges. 

Our baseline algorithms are SAT \citep{goh2022_SAT}, SCNN \citep{Yang_2022_SCNN}, and Simplicial MPNN \citep{bodnar_2021_weisfeiler}. 
We will follow the batch training procedure seen in SAT \citep{goh2022_SAT}: 40 trajectories are batched together to formulate a multidimensional simplicial data/
Each experiment will consist of 5000 iterations for all tested algorithms.
For SAT, SCNN, and Bi-SCNN, we will be running two different architectural settings: 2 simplicial layers and 3 simplicial layers. 
As suggested by \citep{goh2022_SAT}, all the simplicial layers are followed by a mean pooling layer, then 2 MLP readout layers with 30 channels, and finally a softmax layer. 
For each architectural setting, there will be three different activation functions: identity (id) function, leaky ReLu (LR) function, and tanh function. 
The SAT suggests the choices of Id and Tanh functions because odd activation functions lead to orientation-equivariant output, which is shown to perform well on a synthetic ocean drifter dataset \citep{goh2022_SAT}. 
The classification accuracy is shown in Table~\ref{ocean_accuracy} and the time to complete the experiments is recorded in Table~\ref{ocean_time}. 
The number of trainable parameters of each setting is shown in Table~\ref{ocean_drifter_parameter_size}. 
All the tested algorithms will have 40 simplicial filters except the MPNN.
Again, due to the fact that an MPNN with a similar hyperparameter setting to the SCNN surpassed the capability of our hardware.
For the classification accuracy of the MPNN, we can refer to the original paper \citep{bodnar_2021_weisfeiler} as it implemented a 4-layer MPNN with batch size 64 and filter size 64 on this exact dataset for the three tested activation functions in this experiment. 
We will a 2-layer MPNN with 1 filter only to measure the runtime of the experiment.
 
In the Ocean Drifter Dataset, we can observe again that the Bi-SCNN has similar classification accuracy as the SCNN and MPNN for all three choices of action functions. 
Considering the fact that Bi-SCNN has the smallest parameter size among Bi-SCNN, SCNN, and MPNN, it is safe to say that the Bi-SCNN is more efficient. 
As for SAT, the reason Bi-SCNN is capable of consistently outperforming SAT is because the Bi-SCNN utilizes the Hodge decomposition: Bi-SCNN optimizes parameters for upper and lower adjacencies separately but the SAT uses the standard Hodge Laplacian. 
Additionally, the processing of the harmonic component of the input simplicial features is absent in SAT, making its simplicial embedding less representative than convolution-based algorithms. 

One may notice that the run time reduction of Bi-SCNN is not as prominent in the ocean drifter complex as in the citation complex. 
This is because when we recorded the run time, the global pooling and the 2 MLP layers also contributed to the time it took to run the simplicial layers. 
The Bi-SCNN by itself does not optimize the global pooling and the 2 MLP layers, so these layers will perform indifferently among Bi-SCNN, SAT, and SCNN. 
Still, Bi-SCNN is the fastest among Bi-SCNN, SAT, SCNN, and MPNN when all three are compared under the same number of simplicial layers. 

\begin{table}[htbp]
    \centering
    \begin{tabular}{c c} 
    \toprule
    Model & Trainable Parameters\\
    \midrule
    SAT-2 & 5762 \\
    SAT-3 & 6722 \\
    SCNN-2 & 4712 \\
    SCNN-3 & 10142\\
    Bi-SCNN-2 & 3750\\
    Bi-SCNN-3 & 6450\\
    MPNN-2 (ours) & 167337644 \\
    MPNN \citep{bodnar_2021_weisfeiler} & OOM \\
    \bottomrule
    \end{tabular}
    \caption{The number of trainable parameters for each model on the Ocean Drifter dataset.}  
    \label{ocean_drifter_parameter_size}
\end{table}


\section{Conclusion}
\label{sec_conclusion}
The Bi-SCNN was developed from a combination of simplicial convolution with weighted binary-sign propagation, effectively incorporating the upper and lower adjacencies to capture high-order structured simplicial features. 
The nonlinearity of Bi-SCNN is naturally introduced by feature normalization and binarization, allowing time-efficient operation of the Bi-SCNN.
Bi-SCNN outperforms previously proposed SNN algorithms on citation and ocean drifter complexes in terms of time efficiency, while maintaining comparable prediction accuracy.

\end{document}